\pdfoutput=1

\documentclass[11pt]{article}
\usepackage{authblk}
\usepackage[acceptedwitha]{tacl2021v1}

\usepackage{times}
\usepackage{latexsym}
\usepackage{cjhebrew}

\usepackage[T5,T1]{fontenc}

\usepackage[utf8]{inputenc}
\DeclareTextSymbolDefault{\ohorn}{T5}
\DeclareTextSymbolDefault{\uhorn}{T5}
\newcommand{\name}{MultiBLiMP 1.0\xspace}
\newcommand{\namenoV}{MultiBLiMP\xspace}

\newcommand{\nlmax}{\mbox{101}~}

\usepackage{microtype}
\usepackage{inconsolata}
\usepackage{graphicx}
\usepackage{booktabs}
\usepackage{multicol}
\usepackage{multirow}

\usepackage{tikz-dependency}
\usepackage{array}
\usepackage{fancyvrb}
\usepackage{xcolor}
\usepackage{colortbl}
\usepackage{arydshln}
\usepackage{tipa}
\usepackage{amsmath}
\usepackage{cleveref}
\usepackage{enumitem}
\usepackage{comment}
\usepackage{bbold}
\usepackage{xspace}
\usepackage{makecell}

\usepackage{scalerel,xparse}
\usepackage{stackengine}

\Crefname{figure}{{Fig.}}{{Figs.}}
\crefname{section}{§}{§§}
\Crefname{section}{§}{§§}
\Crefname{appendix}{{App.}}{{Apps.}}

\newcommand{\new}[1]{\textcolor{black}{#1}}

\definecolor{customred}{HTML}{C10B14}
\definecolor{customgreen}{HTML}{00883A}

\title{Multi{BL}i{MP} 1.0:\\ A Massively Multilingual Benchmark of Linguistic Minimal Pairs}

\makeatletter
\renewcommand\AB@affilsepx{\hspace{1em} \protect\Affilfont}
\makeatother

\author[1]{\bf Jaap Jumelet}
\author[2]{\bf Leonie Weissweiler}
\author[2]{\bf Joakim Nivre}
\author[1]{\bf Arianna Bisazza}

\affil[1]{University of Groningen} 
\affil[2]{Uppsala University
 \protect\\ {\small\texttt{j.w.d.jumelet@rug.nl, \{leonie.weissweiler,joakim.nivre\}@lingfil.uu.se, a.bisazza@rug.nl}}} 

\begin{document}
\maketitle
\begin{abstract}
We introduce \name{}, a massively multilingual benchmark of linguistic minimal pairs, covering \nlmax languages and 2 types of subject-verb agreement, containing more than 128,000 minimal pairs.
Our minimal pairs are created using a fully automated pipeline, leveraging the large-scale linguistic resources of Universal Dependencies and UniMorph.
\name{} evaluates abilities of LLMs at an unprecedented multilingual scale, and highlights the shortcomings of the current state-of-the-art in modelling low-resource languages.
\footnote{Code: {\tt\href{https://github.com/jumelet/multiblimp}{github.com/jumelet/multiblimp}}\\Data: {\tt\href{https://huggingface.co/datasets/jumelet/multiblimp}{huggingface.co/datasets/jumelet/multiblimp}}}
\end{abstract}

\section{Introduction}

Large language models (LLMs) are often trained on highly multilingual corpora, which enable users to interact with them in a wide range of languages \citep{llama3, ustun-etal-2024-aya}.
Multilingual evaluation of LLMs, however, has mostly focused on their \textit{functional linguistic competence} through tasks requiring world knowledge and language understanding \citep{singh-etal-2024-aya,DBLP:journals/corr/abs-2412-03304}, while little work has assessed their \textit{formal linguistic competence}, i.e. the ``knowledge of rules and statistical regularities of a language'' \citep{MAHOWALD2024517}.
%
In multilingual contexts, the latter is commonly approximated intrinsically through perplexity \citep{goldfish} or extrinsically through performance in generative tasks such as translation or summarization \citep{dang2024aya}. 
These approaches do not truly disentangle formal from functional competence. Moreover, they are very coarse-grained and do not inform us on which specific constructions a model does (not) master.


\begin{figure*}[t!]
    \centering
    \includegraphics[width=\textwidth]{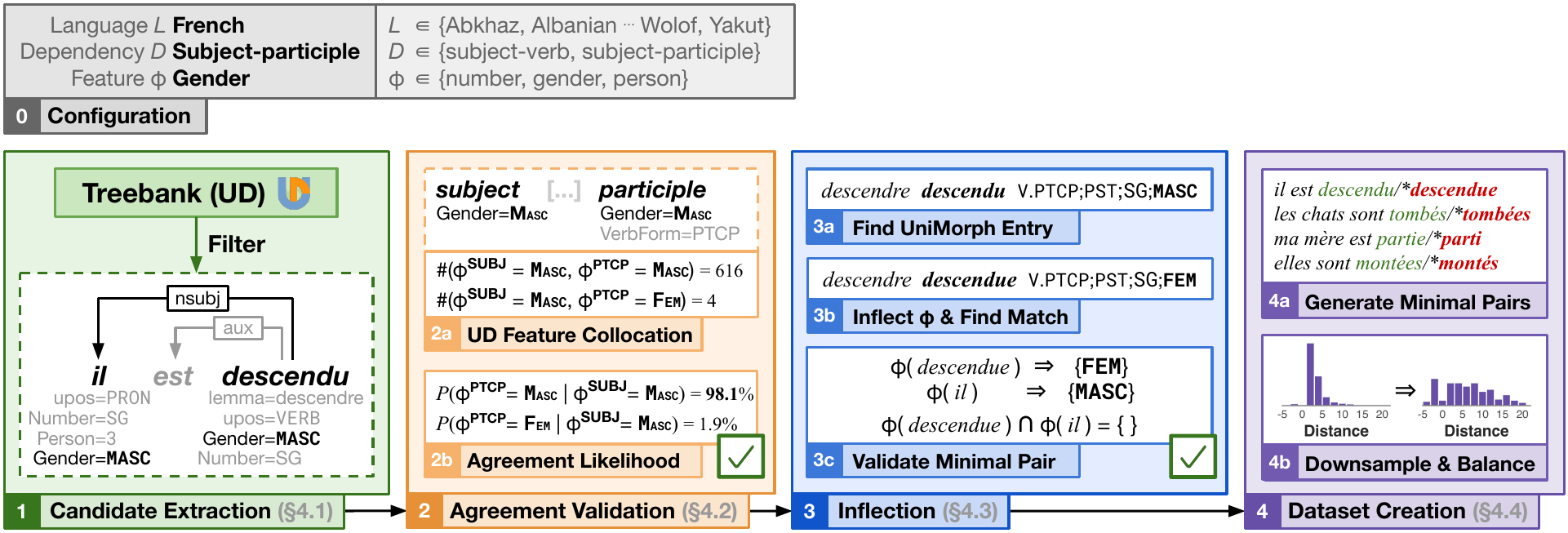}
    \caption{Pipeline of the minimal pair creation procedure of \name{} }
    \label{fig:pipeline}
\end{figure*}

Both shortcomings are addressed by the design of targeted syntactic evaluation benchmarks, typically structured as pairs of grammatical/ungrammatical sentences differing by a single syntactic aspect \citep{linzen-etal-2016-assessing, warstadt-etal-2020-blimp-benchmark}, where a formally competent LM is expected to assign higher probability to the grammatical version.
Such datasets, however, exist only for English and a few other, mostly high-resource languages \citep{gulordava-etal-2018-colorless,mueller-etal-2020-cross,taktasheva-etal-2024-rublimp}.


To accelerate progress in this direction, we introduce \textbf{\name}, a massively multilingual benchmark of linguistic minimal pairs 
covering two types of subject-verb agreement (\new{subject-finite-verb and subject-participle}) for number, person, and gender; created automatically using two large-scale linguistic resources: Universal Dependencies \cite[UD,][]{nivre-etal-2016-universal,nivre-etal-2020-universal,de-marneffe-etal-2021-universal} and UniMorph \citep{batsuren-etal-2022-unimorph}.
Multi\-BLiMP is not only a \textit{benchmark}, but also a \textit{pipeline} for the automatic creation of highly multilingual benchmarks (\Cref{fig:pipeline}), which can scale to many more linguistic phenomena.
%

Taking \new{subject-finite-verb\footnote{Referred to in the following as subject-verb agreement for simplicity.} and subject-participle} agreement as a use case, we present a first version of the benchmark including 
more than 128,000 minimal pairs across \nlmax{} languages. 
We use this to evaluate 42 LLMs, finding that linguistic competence is strongly driven by model size and language frequency in training data (\Cref{fig:freq_acc}), is acquired during pre-training, and can deteriorate as a result of post-training.

Besides its practical relevance for LLM evaluation, MultiBLiMP enables the study of the learnability of specific linguistic constructions cross-lingually in a unified framework, at a much larger scale than currently possible \cite{mueller-etal-2020-cross}, with future applications in LM interpretability \citep{brinkmann2025largelanguagemodelsshare} and quantitative typology \citep{levshina2023we,baylor-etal-2024-multilingual}.







\section{Background}


\subsection{Targeted Syntactic Evaluation}
The syntactic abilities of language models have commonly been evaluated using syntactic minimal pairs: sentence pairs that are altered in a minimal way to create a specific grammaticality violation.
By comparing a model's probability judgment of the grammatical sentence to that of the ungrammatical one, we can assess whether it has acquired a notion of the underlying phenomenon (we provide a technical description in \S\ref{sec:metrics}).

This approach to test language models was introduced by \citet{linzen-etal-2016-assessing}, focusing on English subject-verb number agreement. 
Specifically, they extracted sentences from Wikipedia and flipped the number of the verb without detecting the subject, finding that models at that time struggled considerably with increased distance between the subject and the verb, as well as intervening nouns and other attractors. 
This work was extended by \citet{marvin-linzen-2018-targeted}, who generated sentences with context-free grammars, enabling them to increase the difficulty of agreement across phrases and clauses. 
BLiMP \citep{warstadt-etal-2020-blimp-benchmark} extended this approach to a wider range of linguistic phenomena, generating minimal pairs with templates for many linguistic phenomena by sampling from a vocabulary. 
\citet{hu-etal-2020-systematic} and \citet{gauthier-etal-2020-syntaxgym} both present similar large-scale collections of syntactic evaluation for English.

Rather than relying on minimal pairs, \citet{pratapa-etal-2021-evaluating} develop a method to assess the morphosyntactic well-formedness of model outputs based on its dependency parse and a set of rules automatically extracted from parsed corpora. 
Their approach is complementary to ours as it focuses on the well-formedness of a machine-generate text as opposed to measuring the intrinsic syntactic abilities of a language model on controlled stimuli. 
However, the requirement of a well-performing parser in each language of interest considerably limits the applicability of their method to a wide set of languages. 

\paragraph{Monolingual Non-English Benchmarks}
Minimal pair benchmarks have recently been developed for various other languages, using different methods. 
Some use existing annotated sentences from UD, some craft templates, and some others are built by translating an English dataset.

For Chinese, \citet[][CLiMP]{xiang-etal-2021-climp} generate minimal pairs with templates adapted from BLiMP, while \citet[][SLING]{song-etal-2022-sling} edit a constituency treebank with templates and then verify the results with human annotators. 
\citet[][ZhoBLiMP]{DBLP:journals/corr/abs-2411-06096} employ a mixture of adapting paradigms from English BLiMP, and extracting Chinese-specific paradigms from syntax journals and textbooks.
\citet[][JBLiMP]{someya-oseki-2023-jblimp} extract minimal pairs from journal articles on Japanese syntax.
\citet[][BLiMP-NL]{suijkerbuijk2024} create ten sentence pairs per paradigm by hand for Dutch, as well as 90 more sentences generated by ChatGPT and then hand-checked. 
\citet[][LINDSEA]{DBLP:journals/corr/abs-2309-06085} present minimal pairs for Tamil and Indonesian which are created manually by linguists and validated by native speakers.
\citet{kryvosheieva-levy-2025-controlled} create minimal pairs specifically for Swahili noun class agreement, Hindi split ergativity and Basque verb agreement, each chosen for their complexity.

Closer to our approach, \citet[][RuBLiMP]{taktasheva-etal-2024-rublimp} create pairs from UD-parsed sentences using perturbation rules. 
Our approach too leverages UD to identify syntactic relations, which we extend multilingually by setting up a language-agnostic minimal pair creation procedure (\S\ref{sec:pipeline}).
%


\paragraph{Multilingual}
Two multilingual minimal pair benchmarks are known to us. 
The largest, CLAMS \citep{mueller-etal-2020-cross}, creates pairs for English, German, Hebrew, Italian, and Russian for subject-verb agreement by having native speakers translate the sentences from \citet{marvin-linzen-2018-targeted} where applicable. 
\citet{gulordava-etal-2018-colorless} similarly cover English, Hebrew, Italian, and Russian for various types of agreement using sentences from UD treebanks, in which content words are swapped out with others of the same part-of-speech to test the models' reliance on semantics.
Two other notable related efforts present a multilingual benchmark for linguistic acceptability---for which a model is fine-tuned to predict the grammaticality of a single sentence instead of judging a pair \citep{warstadt-etal-2019-neural}---ScaLA, which is part of ScandEval \citep{nielsen-2023-scandeval}, covering 5 Scandinavian languages, and MELA \citep{zhang-etal-2024-mela}, covering a diverse set of 10 languages.

We see \name as complementary to all these efforts.
While language-specific benchmarks can provide tremendous depth on the syntactic system of a language, it will prove challenging to scale such efforts to a wide variety of languages.
Our approach, on the other hand, focuses on `wide coverage' and can therefore be of use for both the evaluation of highly multilingual LLMs and for quantitative typological studies. 


\subsection{Representation Sharing in LLMs}

Various works have evaluated the degree to which multilingual models make use of shared syntactic representations across languages.
Early studies analyzed mBERT \citep{devlin-etal-2019-bert} and smaller-scale LSTM models trained on a few languages, finding mixed evidence of sharing \citep{pires-etal-2019-multilingual,mueller-etal-2020-cross,dhar-bisazza-2021-understanding,chi-etal-2020-finding, choenni-shutova-2022-investigating}.
Focusing on modern-scale LLMs, \citet{wendler-etal-2024-llamas} find that 
models trained on English-dominated corpora use English to some degree as an internal `pivot language' to solve tasks in other languages. 
\citet{brinkmann2025largelanguagemodelsshare} find significant overlap between the representations for number, gender, tense and other morphosyntactic features.
Similarly, \citet{stanczak-etal-2022-neurons} find overlap in the presentations of morphosyntactic features, particularly for grammatical number. 
\citet{ferrando-costa-jussa-2024-similarity} show the similar structures of circuits for subject-verb agreement in English and Spanish. 
The emergence of this kind of interpretability findings highlights the exciting opportunities presented by \name{} for massively multilingual interpretability.

\section{Linguistic Resources}
We briefly describe here the key linguistic resources that we use to create MultiBLiMP: Universal Dependencies and UniMorph.

\subsection{Universal Dependencies}
\label{sec:treebanks}

Universal Dependencies \cite{nivre-etal-2016-universal,nivre-etal-2020-universal,de-marneffe-etal-2021-universal} is a multilingual treebank collection, containing 296 treebanks for 168 languages.
Besides syntactic dependency relations, UD also contains rich morphological feature annotations that we leverage in our pipeline. 
An example of a UD-annotated sentence can be found in \Cref{fig:pipeline}, step 1. We use UD version 2.15 for the 1.0 release of MultiBLiMP \cite{ud_v215}.

We exclude treebanks of spoken language, historical varieties, as well as certain genres, e.g. technical manuals. 
Generally, we only exclude treebanks for low-resource languages where absolutely necessary, while for high-resource languages, we also exclude very small treebanks that will be of little additional benefit, but may introduce annotation inconsistencies. 
A full list of the excluded treebanks with reasons for exclusion can be found in \Cref{app:excl_treebanks}.
We then consider the union of all remaining treebanks for each language.


\subsection{UniMorph}


UniMorph \citep[UM, \texttt{v4.0},][]{batsuren-etal-2022-unimorph} is a multilingual collection of morphological feature annotations, which define word-level features such as \textsc{number}, \textsc{mood}, or \textsc{gender} for nouns, verbs, and adjectives.
It covers 183 languages in total, 81 of which are also covered by UD.
Similarly to UD, it uses a universal set of features \citep{sylak-glassman-etal-2015-language}, making it highly suitable for our goal of defining a language-independent inflection mechanism to create minimal pairs.
UM entries contain the lemma and morphological features of a word form, making it possible to efficiently disambiguate and create inflections.
For example, the English word \textit{saw} is represented with the following four entries, expressing its various meanings:

\begin{small}
\begin{Verbatim}[commandchars=\\\{\}]
\textcolor{gray}{lemma}  \textcolor{gray}{form}   \textcolor{gray}{features}
see    saw    V;PST
saw    saw    N;SG
saw    saw    V;NFIN;IMP+SBJV
saw    saw    V;PRS;3;IND;PL    
\end{Verbatim}
\end{small}
\vspace{-.3em}
\noindent
Note that features can be left implicit: the first entry of past tense \textit{see} does not specify \textsc{person} or \textsc{number}, indicating that this form covers all values of those features.
We preprocess UM features to be compatible with UD feature values (\texttt{SG} $\rightarrow$ \textit{Sing}, \texttt{PRS} $\rightarrow$ \textit{Pres}, etc.), and transliterate languages that use different scripts from those in UD.

\paragraph{UD Features}
To broaden the number of languages for which we can create inflections, we leverage the morphological feature annotations that are present in UD, extending the approach of UDLexicons \citep{sagot-2018-multilingual}---which covers 38 languages---to the 142 languages in UD that contain annotations.
For each language, we extract the $\langle\textit{lemma},~\textit{form},~\textit{features}\rangle$ triplets for each token in the treebank.
To ensure the quality of these triplets, we incorporate their frequency to filter out potential annotation errors.
We make the assumption that a $\langle\textit{lemma},~\textit{features}\rangle$ tuple should map onto a single \textit{form}; if the tuple yields multiple forms, it is likely that a feature has been annotated erroneously.\footnote{For example, $\langle$\textit{walk}, \textsc{Tense=Pres;} \textsc{Person=3;} \textsc{Mood=Ind;} \textsc{Number=Sing}$\rangle$ should map only to \textit{walks}; if it would also yield \textit{walk} it indicates an annotation error.
Since such annotation errors are infrequent, we select the most frequent item instead.
}
In such cases, we discard all entries occurring three times less than the most frequent entry.
This procedure results in 4.2M unique triplets for 142 languages, 61 of which are not covered by UM.


\section{Pipeline}\label{sec:pipeline}
Our pipeline for creating minimal pairs consists of four stages.
First, we \textbf{extract} suitable candidates for a phenomenon using dependency parse trees (\S\ref{sec:extraction}).
This allows us to determine the key items of interest for a phenomenon, like the subject and verb in subject-verb agreement.
Next, we \textbf{validate} whether the agreement phenomenon is present in that language based on UD collocation statistics (\S\ref{sec:agreement}).
We then \textbf{inflect} one of the words with respect to a particular \textit{feature} (e.g. \textsc{number}) to create an agreement violation (\S\ref{sec:inflect}).
By replacing the item with this inflected form, we create the ungrammatical counterpart for the minimal pair.
Finally, we create a minimal pair dataset for each language, \textbf{balancing} the morphological features and number of pairs (\S\ref{sec:balance}).
We now explain this procedure in more detail; a graphical overview is provided in Figure~\ref{fig:pipeline}.

\subsection{Candidate Extraction}
\label{sec:extraction}
The 167 UD languages form the initial set of candidate languages we consider for our benchmark.
After having excluded some treebanks as described in \S\ref{sec:treebanks}, we also set various sentence-level constraints. Specifically, we discard sentences containing \texttt{reparandum} dependencies or a \texttt{Style}, \texttt{Foreign} or \texttt{Typo} feature, which typically signal sentences that are malformed in some way.

\paragraph{Filters}\label{sec:filters}
To extract candidates for a particular linguistic phenomenon, we define filters based on dependency edges, part-of-speech tags and morphological features.
In this paper, we restrict our attention to various forms of subject-verb and subject-participle agreement.
The filters for this phenomenon are defined as follows:
\begin{center} 
\begin{dependency}
    \begin{deptext}[column sep=1.5em]
        \textit{subject} \& \textcolor{customred}{...} \& \textcolor{gray}{\textit{aux}} \& \textit{verb} \& \textcolor{customred}{...}\\
        \textsc{noun} \& \& \textcolor{gray}{\textsc{aux}} \& \textsc{verb} \& \\ 
    \end{deptext}
    \depedge[edge height=1.7em]{4}{1}{nsubj}
    \depedge[edge height=1em,edge style={customred, thick}, label style={text=customred,draw=customred}]{1}{2}{conj}
    \depedge[edge style={customred, thick}, label style={text=customred,draw=customred, inner ysep=3.5ex}, edge height=2em]{4}{5}{\makecell{expl\\csubj:outer\\nsubj:outer}}
    \depedge[edge height=1em,edge style={gray, thick}, label style={text=gray,draw=gray}]{4}{3}{aux/cop}
\end{dependency} 
\end{center}
That is, we look for \texttt{nsubj} dependency edges between a \textsc{noun} and a \textsc{verb} and collect any auxiliary verb connected to it.
If the \textsc{verb} or one of its auxiliaries has the feature \textsc{VerbForm=Finite}, it is considered as a candidate for subject-verb agreement.
For finding subject-participle agreement relations, we place an additional constraint on the morphological features of the verb of \textsc{VerbForm=Part}.
We drop subjects containing conjunctions, since these may have conflicting feature values.
Furthermore, we drop verbs that have an outgoing \texttt{expl}, \texttt{csubj:outer}, or \texttt{nsubj:outer} dependency, since these indicate clausal subject constructions that may invalidate our minimal pair formation methodology.


\subsection{Agreement Validation}\label{sec:agreement}
Determining whether a particular phenomenon exists in a given language, and how it can be expressed as a minimal pair, is challenging.
We address this with a data-driven procedure that computes the probability of a language containing a phenomenon, based on UD and UM annotations.\footnote{An alternative to approaching this from a data-driven perspective would be to use typological databases like Grambank \citep{grambank_release} and WALS \citep{wals}, but their discrete nature and framing agreement in terms of being \textit{possible} and not \textit{compulsory} \citep{baylor-etal-2023-past} make them unsuitable for our goals.}
This acts as a final filter on the minimal pairs, ensuring that 
our perturbed sentences are actually ungrammatical in a given language.\footnote{In addition to this data-driven procedure, we also conduct extensive spot checks to ensure our procedure yields valid minimal pairs.}

\paragraph{Agreement}
The agreement phenomena we focus on can be turned into a minimal pair by changing the agreement feature on a word and re-inflecting it:
\begin{center}
\stackon[1ex]{\small\textcolor{gray}{\hspace{0.85cm}\texttt{N;PL}\hspace{0.24cm}\texttt{V;\textbf{PL}}}}{\normalsize\textit{The~~boys~~walk}}
\stackon[1ex]{\textcolor{white}{\texttt{X}}}{~\large\textit{$\Rightarrow$}~}
\stackon[1ex]{\small\textcolor{gray}{\hspace{0.85cm}\texttt{N;PL}\hspace{0.3cm}\texttt{V;\textbf{SG}}}}{\normalsize\textit{$^*$The~~boys~~walks}}
\end{center}
Evaluating agreement in this way requires that the inflected form is ungrammatical in this context. 
For example, if we were to inflect a past tense verb for \textsc{number} in English, the inflected form would remain the same.
Another example is Turkish, where number is marked \textit{optionally} on the verb: 
\begin{center}
\stackon[1ex]{\small\textcolor{gray}{\texttt{N;PL}\hspace{0.8cm}\texttt{V;\textbf{PL}}}}{\normalsize\textit{Oğlanlar~~yürüyorlar}}
\stackon[1ex]{\textcolor{white}{\texttt{X}}}{~\large\textit{$\Rightarrow$}~}
\stackon[1ex]{\small\textcolor{gray}{\texttt{N;PL}\hspace{0.8cm}\texttt{V;\textbf{SG}}}}{\normalsize\textit{Oğlanlar~~yürüyor}}
\end{center}
Both these constructions are correct, and the construction with the singular verb is even more frequent. 
A model provided with this minimal pair would likely assign a higher probability to the sentence with the singular verb and be penalized for it, resulting in an incorrect assessment of the model's grammatical abilities.

\paragraph{Requirements}
Based on these observations, we define two requirements to determine if a language contains agreement for some feature $\phi$ \citep{baker2008syntax}:
\begin{enumerate}
    \item[R1:] $\phi$ must be specified for both elements of the agreement relation, either lexically inherent (\textit{la \underline{fille} est tombé\underline{e}};  \textit{the girl fell}) or morphologically realized (\textit{les garçon\underline{s} sont tombé\underline{s}}; \textit{the boys fell}).
    This way, for example, we rule out subject-verb gender agreement in English, since gender is not specified on verbs.
    We easily validated this using UniMorph and UD by checking whether both subjects and verbs are annotated for the feature.
    \item[R2:] Re-inflecting one of the elements with a different value for $\phi$ for one of the elements must result in a grammatical violation. This requirement is more challenging to test.
\end{enumerate}

\paragraph{Collocations}
We approach Requirement 2 from a collocational perspective---a similar approach was used by \citet{chaudhary-etal-2020-automatic} for grammatical rule detection from treebanks.
In a language having strict subject-verb agreement for $\phi$, we expect very high co-occurrence between nouns and verbs of the same feature value, and low co-occurrence between nouns and verbs of contrasting values.
This can be expressed as a conditional probability: 
\begin{align}\label{eq:collocation}
    P_{agr}(\phi^v=x|\phi^n=x, \textup{WO}) \approx 1
\end{align}

\noindent where $\phi^v$ and $\phi^n$ denote the feature of verb and noun, $x$ a specific feature value (e.g. \textsc{sg} or \textsc{pl}), and WO the subject-verb word order (SV/VS).
If this probability is close to 1, it implies that the feature of the noun co-occurs highly with that of the verb, and the probability of the verb having a contrasting feature value is close to 0.


\begin{table}[t!]
    \centering
    \small
    \begin{tabular}{ccc rr}
        & & & \multicolumn{2}{c}{$P_{agr}(\phi^v|\phi^n,\text{WO})$} \\\cmidrule(lr){4-5}
        $\phi^n$ & WO & $\phi^v$ & Dutch & Turkish \\
        \midrule
\multirow{2}{*}{PL} & \multirow{2}{*}{SV} & PL & \textbf{0.989} &  0.161 \\
 & & SG &  0.011  &  0.839 \\\arrayrulecolor[rgb]{0.753,0.753,0.753}\midrule\arrayrulecolor[rgb]{0.0, 0.0, 0.0}
\multirow{2}{*}{PL} & \multirow{2}{*}{VS} & PL & \textbf{0.990} &  0.000 \\
 & & SG & 0.010  & {1.000} \\\arrayrulecolor[rgb]{0.753,0.753,0.753}\midrule\arrayrulecolor[rgb]{0.0, 0.0, 0.0}
\multirow{2}{*}{SG} & \multirow{2}{*}{SV} & PL & 0.012  &  0.025 \\
 & & SG &  \textbf{0.988} & \textbf{0.975} \\\arrayrulecolor[rgb]{0.753,0.753,0.753}\midrule\arrayrulecolor[rgb]{0.0, 0.0, 0.0}
\multirow{2}{*}{SG} & \multirow{2}{*}{VS} & PL & 0.008  &  0.018 \\
 & & SG &  \textbf{0.992} & \textbf{0.982} \\
        \bottomrule
    \end{tabular}
    \caption{Dutch and Turkish agreement probabilities for subject-verb number agreement.
    Significant agreement is denoted in \textbf{boldface}. WO stands for word order, either subject verb (SV) or verb subject (VS).
    }
    \label{tab:agr_example}
\end{table}

\begin{table*}[t]
\centering
\footnotesize
\begin{tabular}{lccccccccc}
\midrule\arrayrulecolor[rgb]{0.753,0.753,0.753}
\textsc{Number}	&	\multicolumn{2}{c}{\textsc{SG}}	&	\multicolumn{2}{c}{\textsc{PL}}	&	\multicolumn{2}{c}{\textsc{DU}}	&	\multicolumn{3}{c}{Total}\\
\cmidrule(lr){2-3} \cmidrule(lr){4-5} \cmidrule(lr){6-7} \cmidrule(lr){8-10}
& \textsc{sv} & \textsc{vs} & \textsc{sv} & \textsc{vs} & \textsc{sv} & \textsc{vs} & \textsc{sv} & \textsc{vs} & \textsc{both} \\
{S-Verb}	&	\cellcolor[rgb]{0.541,0.809,0.533}$88$ {\scriptsize$(74)$}	&	\cellcolor[rgb]{0.585,0.829,0.570}$82$ {\scriptsize$(63)$}	&	\cellcolor[rgb]{0.644,0.856,0.620}$73$ {\scriptsize$(58)$}	&	\cellcolor[rgb]{0.732,0.894,0.706}$58$ {\scriptsize$(43)$}	&	\cellcolor[rgb]{0.944,0.979,0.932}$8$ {\scriptsize$(3)$}	&	\cellcolor[rgb]{0.958,0.984,0.948}$3$ {\scriptsize$(2)$}	&	\cellcolor[rgb]{0.535,0.806,0.529}$89$ {\scriptsize$(79)$}	&	\cellcolor[rgb]{0.568,0.822,0.556}$84$ {\scriptsize$(65)$}	&	\cellcolor[rgb]{0.530,0.804,0.524}$90$ {\scriptsize$(80)$}	\\
{S-Participle}	&	\cellcolor[rgb]{0.850,0.941,0.827}$35$ {\scriptsize$(33)$}	&	\cellcolor[rgb]{0.883,0.955,0.862}$28$ {\scriptsize$(21)$}	&	\cellcolor[rgb]{0.902,0.962,0.883}$23$ {\scriptsize$(18)$}	&	\cellcolor[rgb]{0.913,0.967,0.896}$19$ {\scriptsize$(13)$}	&	\cellcolor[rgb]{0.958,0.984,0.948}$3$ {\scriptsize$(0)$}	&	\cellcolor[rgb]{0.964,0.987,0.956}$1$ {\scriptsize$(0)$}	&	\cellcolor[rgb]{0.850,0.941,0.827}$35$ {\scriptsize$(33)$}	&	\cellcolor[rgb]{0.872,0.950,0.850}$30$ {\scriptsize$(22)$}	&	\cellcolor[rgb]{0.850,0.941,0.827}$35$ {\scriptsize$(33)$}	\\\arrayrulecolor[rgb]{0.753,0.753,0.753}\midrule\arrayrulecolor[rgb]{0.753,0.753,0.753}

\textsc{Gender}	&	\multicolumn{2}{c}{\textsc{MASC}}	&	\multicolumn{2}{c}{\textsc{NEUT}}	&	\multicolumn{2}{c}{\textsc{FEM}}	&	\multicolumn{3}{c}{Total}\\
\cmidrule(lr){2-3} \cmidrule(lr){4-5} \cmidrule(lr){6-7} \cmidrule(lr){8-10}
& \textsc{sv} & \textsc{vs} & \textsc{sv} & \textsc{vs} & \textsc{sv} & \textsc{vs} & \textsc{sv} & \textsc{vs} & \textsc{both} \\
{S-Verb}	&	\cellcolor[rgb]{0.868,0.949,0.846}$31$ {\scriptsize$(16)$}	&	\cellcolor[rgb]{0.887,0.956,0.866}$27$ {\scriptsize$(16)$}	&	\cellcolor[rgb]{0.931,0.974,0.917}$13$ {\scriptsize$(7)$}	&	\cellcolor[rgb]{0.931,0.974,0.917}$13$ {\scriptsize$(6)$}	&	\cellcolor[rgb]{0.872,0.950,0.850}$30$ {\scriptsize$(13)$}	&	\cellcolor[rgb]{0.898,0.961,0.878}$25$ {\scriptsize$(11)$}	&	\cellcolor[rgb]{0.853,0.943,0.831}$34$ {\scriptsize$(19)$}	&	\cellcolor[rgb]{0.872,0.950,0.850}$30$ {\scriptsize$(16)$}	&	\cellcolor[rgb]{0.853,0.943,0.831}$34$ {\scriptsize$(22)$}	\\
{S-Participle}	&	\cellcolor[rgb]{0.890,0.958,0.870}$26$ {\scriptsize$(25)$}	&	\cellcolor[rgb]{0.898,0.961,0.878}$25$ {\scriptsize$(20)$}	&	\cellcolor[rgb]{0.935,0.975,0.922}$11$ {\scriptsize$(8)$}	&	\cellcolor[rgb]{0.935,0.975,0.922}$11$ {\scriptsize$(8)$}	&	\cellcolor[rgb]{0.911,0.966,0.894}$20$ {\scriptsize$(17)$}	&	\cellcolor[rgb]{0.911,0.966,0.894}$20$ {\scriptsize$(12)$}	&	\cellcolor[rgb]{0.890,0.958,0.870}$26$ {\scriptsize$(25)$}	&	\cellcolor[rgb]{0.890,0.958,0.870}$26$ {\scriptsize$(21)$}	&	\cellcolor[rgb]{0.890,0.958,0.870}$26$ {\scriptsize$(25)$}	\\\arrayrulecolor[rgb]{0.753,0.753,0.753}\midrule\arrayrulecolor[rgb]{0.753,0.753,0.753}

\textsc{Person}	&	\multicolumn{2}{c}{\textsc{1}}	&	\multicolumn{2}{c}{\textsc{2}}	&	\multicolumn{2}{c}{\textsc{3}}	&	\multicolumn{3}{c}{Total}\\
\cmidrule(lr){2-3} \cmidrule(lr){4-5} \cmidrule(lr){6-7} \cmidrule(lr){8-10}
& \textsc{sv} & \textsc{vs} & \textsc{sv} & \textsc{vs} & \textsc{sv} & \textsc{vs} & \textsc{sv} & \textsc{vs} & \textsc{both} \\
{S-Verb}	&	\cellcolor[rgb]{0.658,0.862,0.633}$71$ {\scriptsize$(42)$}	&	\cellcolor[rgb]{0.751,0.901,0.724}$55$ {\scriptsize$(25)$}	&	\cellcolor[rgb]{0.704,0.882,0.679}$63$ {\scriptsize$(32)$}	&	\cellcolor[rgb]{0.835,0.936,0.811}$38$ {\scriptsize$(15)$}	&	\cellcolor[rgb]{0.579,0.827,0.565}$83$ {\scriptsize$(62)$}	&	\cellcolor[rgb]{0.667,0.866,0.643}$69$ {\scriptsize$(38)$}	&	\cellcolor[rgb]{0.557,0.816,0.547}$86$ {\scriptsize$(66)$}	&	\cellcolor[rgb]{0.639,0.854,0.615}$74$ {\scriptsize$(44)$}	&	\cellcolor[rgb]{0.552,0.814,0.542}$87$ {\scriptsize$(68)$}	\\
{S-Participle}	&	\cellcolor[rgb]{0.962,0.986,0.953}$2$ {\scriptsize$(1)$}	&	\cellcolor[rgb]{0.966,0.987,0.958}$0$ {\scriptsize$(0)$}	&	\cellcolor[rgb]{0.964,0.987,0.956}$1$ {\scriptsize$(1)$}	&	\cellcolor[rgb]{0.966,0.987,0.958}$0$ {\scriptsize$(0)$}	&	\cellcolor[rgb]{0.962,0.986,0.953}$2$ {\scriptsize$(2)$}	&	\cellcolor[rgb]{0.964,0.987,0.956}$1$ {\scriptsize$(0)$}	&	\cellcolor[rgb]{0.958,0.984,0.948}$3$ {\scriptsize$(2)$}	&	\cellcolor[rgb]{0.964,0.987,0.956}$1$ {\scriptsize$(0)$}	&	\cellcolor[rgb]{0.958,0.984,0.948}$3$ {\scriptsize$(2)$}	\\
\arrayrulecolor{black}\midrule
\textsc{Total} & & & & & & &$100$ {\scriptsize$(85)$}	&	$90$ {\scriptsize$(66)$}	&	$101$ {\scriptsize$(87)$}

\end{tabular}
\caption{
The number of unique languages that yielded at least 10 minimal pairs, for each agreement condition.
The number between brackets denotes the number of languages for which the binomial test of \S\ref{sec:agreement} was significant.
SV and VS denote the subject-verb word order.
}\label{tab:num_pairs}
\end{table*}

Agreement can depend on various extra factors, such as word order. 
For example, person agreement in Dutch depends on this: second person singular verbs in VS order lose their second person ending and coincide with first person (\textit{jij doe\textbf{t}} vs. \textit{doe jij}).
A different example is French, for which subject-participle agreement depends on the auxiliary: participles inflect for \textsc{number} and \textsc{gender} with an \textit{être} auxiliary, but not with an \textit{avoir} auxiliary (\textit{elles sont monté\textbf{es}} vs. \textit{elles ont monté}).
Arabic has flexible subject-verb order and does not inflect for plural in VS order, but only has agreement in SV order for human subjects.
In our pipeline we therefore condition Equation~\ref{eq:collocation} on a binary word order feature, but leave a more fine-grained exploration of agreement conditions open for future work, for instance by extending the data-driven agreement extraction approach of \citet{chaudhary-etal-2020-automatic} with word order and lexical features.

\paragraph{Agreement Conditions}
We compute agreement probabilities for each language, verb type (finite main verb, finite auxiliary or participle), feature value, and word order. 
We refer to this configuration as an \textbf{agreement condition}.
Probabilities for each agreement condition are computed by counting feature co-occurrences based on the morphological annotations in the UD tree, using the dependency relations we extract from UD (\S\ref{sec:filters}).
In case a feature is not annotated, we look it up in UniMorph and, if present, take the value from there.
We only count co-occurrences for cases where a feature is defined for both the subject and the verb (e.g. \textsc{number} is not specified for past tense verbs in English).
Based on the co-occurrence counts of the feature values, we then compute the conditional probability distribution of Equation~\ref{eq:collocation}.

Note that it is possible this way for a language to have strict agreement in only one direction.
We show an example in Table~\ref{tab:agr_example}, providing subject-verb number agreement probabilities for two languages: Dutch, which has strict number agreement for all conditions, and Turkish, where plurality is optionally marked on the verb, but singular verbs are commonly used for both singular and plural subjects. 
Dutch number agreement is marked as significant for both singular and plural, whereas Turkish agreement only demonstrates agreement for singular items.

\paragraph{Significance}
Since our procedure depends on the amount of co-occurrence counts we can extract from UD for a language, we need to determine whether the collocation probability is statistically significant.
For this, we conduct a binomial test to assess whether Equation~\ref{eq:collocation} was significantly greater than the baseline $p_0 = 0.9$.
To rule out agreement, we conduct a binomial test in the opposite direction. 
If neither test is significant, we mark agreement as being \textit{uncertain}. 
For significance we use $\alpha = 0.1$ with Bonferroni correction.
We only proceed to the minimal pair creation step for the agreement conditions that have certain and uncertain agreement. The conditions with no agreement are discarded.

\begin{figure*}[ht!]
    \centering
    \includegraphics[width=\textwidth]{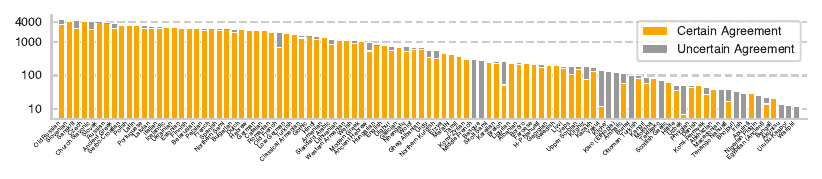}\vspace{-0.5cm}
    \caption{
    Number of minimal pairs per language in MultiBLiMP, split out for certain and uncertain agreement cases using the agreement detection procedure of \S\ref{sec:agreement}. Note the log-scale on the y-axis.
    }
    \label{fig:pair_distribution}
\end{figure*}

\subsection{Inflection}\label{sec:inflect}
To form the ungrammatical counterpart of a sentence extracted from UD, we inflect a specific word to a different feature that makes the sentence ungrammatical.
For example, to create a minimal pair for subject-verb number agreement for \textit{The boy walks}, we re-inflect the verb \textit{walk} with \textsc{number=Plur} to create \textit{$^*$The boy walk}.
While we focus here on \textsc{gender}, \textsc{person}, and \textsc{number}, this can work for any agreement feature annotated on the verb.

Our inflection procedure has two stages.
We start by defining the feature to inflect for, with a candidate word from the filtered treebank of \Cref{sec:extraction}.
The first stage is to find the corresponding lemma and morphological features of this word in the UD and UM databases, based on its form and the features that were annotated in the UD tree.
This may yield multiple candidate lemmas and features, depending on the level of detail of the UD features. 
In the second stage, we then find all matching rows containing \textit{contrasting values} of the feature that we inflect for.
The inflected forms of these matching rows are the inflection candidates that we use to create the minimal pairs.
By considering all opposite values we may create multiple pairs from a single sentence, e.g. \textit{He is [..]}~\textrightarrow~\textit{He are/am [..]}.
%


\subsection{Dataset Balancing}\label{sec:balance}
To level data imbalances, we limit the number of minimal pairs per agreement condition to 100 items. 
We obtain these 100 items using a weighted downsampling procedure that also balances sentence-level features.
The features we incorporate in this procedure are: \textit{subject form}, \textit{verb form}, \textit{subject-verb distance}, \textit{all attractors congruent}, and \textit{all attractors incongruent}.
The last two features indicate whether the intervening material between subject and verb contained items that had a congruent or incongruent feature with respect to the agreement feature that we re-inflect (e.g. \textit{The \textbf{keys} to the \underline{cabinet} \textbf{are}}).
We define the sampling probability $Q$ of a minimal pair as the inverse of the joint probability of its features $P(x)$, assuming feature independence: 
\[P(x) = \prod_i P(x_i)\hspace{1cm} Q(x) = \frac{P(x)^{-1}}{\sum_i P({x}_i)^{-1}}\]
The feature probability $P(x_i)$ is expressed as the relative frequency of the feature. 
By sampling from $Q$ (without replacements), we can balance all features simultaneously. 
This results in a dataset that is as balanced as possible for multiple features, ensuring both lexical diversity and an even spread of easier and harder sentences.

\section{\name}
\label{sec:pairs}

\begin{figure}[t]
    \centering
    \includegraphics[width=\columnwidth]{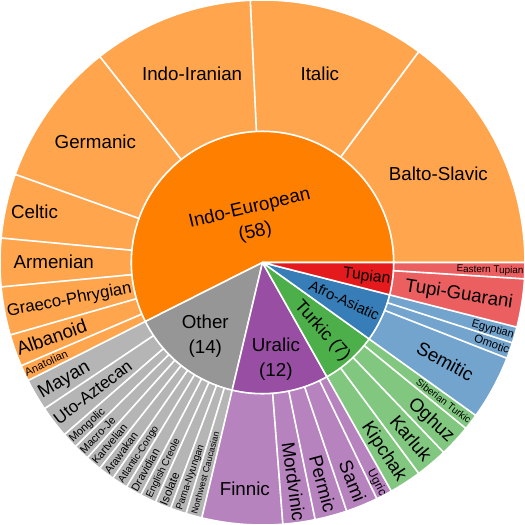}
    \caption{
    Distribution of language families present in \name. See Appendix~\ref{app:lang_distribution} for a detailed version of this figure, including the individual languages.
    }
    \label{fig:fam_distribution}
\end{figure}

We run our minimal pair creation pipeline for two agreement phenomena and three features: subject-verb and subject-participle agreement for \textsc{number}, \textsc{person}, and \textsc{gender}.
Within each condition we create sub-conditions based on the inflected feature (e.g. \texttt{SG} $\rightarrow$ \texttt{PL}, \texttt{2} $\rightarrow$ \texttt{3}, etc.), and the order of subject and verb.
Our procedure results in \textbf{128,321 minimal pairs} across \textbf{\nlmax{} languages}.
The total number of minimal pairs we obtain before balancing is 1.4 million.
We provide an aggregated overview of the number of languages per condition in \Cref{tab:num_pairs}, and a sample of pairs in \Cref{app:pair-examples}.
Subject-verb number agreement is the most common agreement type, with 90 languages, whereas we find subject-participle person agreement for only 3 languages.
We present the number of minimal pairs per language in Figure~\ref{fig:pair_distribution}.
We also provide a detailed breakdown of the language families that are currently covered in \name in Figure~\ref{fig:fam_distribution}.
There is a strong Indo-European (IE) bias present in our minimal pairs, caused both by the over-representation of such languages in UD, as well as our choice to focus on these types of agreement, which tend to be more present in IE languages.
In future iterations of \namenoV, we intend to broaden our coverage by focusing on a more diverse set of phenomena.

\section{LLM Evaluation Setup}
\subsection{Metrics}
\label{sec:metrics}
Following prior work \citep{marvin-linzen-2018-targeted} we measure the grammaticality of the predictions of a model based on the probability it assigns to a grammatical sentence versus an ungrammatical one.\footnote{
The best way to evaluate instruction-tuned LLMs on minimal pairs and grammaticality remains a topic of debate. 
`Meta-linguistic' evaluation assesses grammaticality judgments not based on direct probability comparisons, but instead prompts the model to pick the grammatical sentence.
While \citet{doi:10.1073/pnas.2309583120} found GPT3's abilities lacking using this approach, \citet{doi:10.1073/pnas.2400917121} show that direct evaluation may yield better results, and \citet{song2025languagemodelsfailintrospect} show that meta-linguistic evaluation depends strongly on model size, and outperforms direct evaluation for larger models.}
We choose to evaluate on the sentence-level, and not at the position of the key items, as has been done previously \citep{linzen-etal-2016-assessing,jumelet-hupkes-2018-language}, as this yields the same metric for both subject-verb word orders.
We use two metrics in our experiments.
The accuracy score is based on the number of items for which the model assigns a higher probability to the grammatical sentence ($s^+$):
\vspace{-.5em}
\begin{equation*}
    Acc(\mathcal{M};\mathcal{D}) = \frac{1}{|\mathcal{D}|}\sum_{s\in \mathcal{D}}\mathbb{1}\left[P_\mathcal{M}(s^{+}) > P_{\mathcal{M}}(s^{-})\right]
\end{equation*}
\vspace{-.5em}

\noindent for model $\mathcal{M}$ and minimal pair dataset $\mathcal{D}$.
We also measure the \textit{certainty} of a model's judgment as the log probability difference for the minimal pair:
\vspace{-.5em}
\begin{equation*}
    \Delta(\mathcal{M};\mathcal{D}) = \frac{1}{|\mathcal{D}|}\sum_{s\in \mathcal{D}}\ln P_\mathcal{M}(s^{+}) - \ln P_{\mathcal{M}}(s^{-})
\end{equation*}
\vspace{-.5em}

\subsection{Language Models}\label{sec:models}

We evaluate the following 42 LLMs, accessed through HuggingFace \citep{wolf-etal-2020-transformers}.
For all models, we evaluate both their \texttt{base} version (the model after \textit{pre-training}), and \texttt{chat} version (the model after \textit{post-training}).
\begin{itemize}[leftmargin=*,noitemsep, topsep=2pt]
    \setlength{\itemsep}{0pt}
    \setlength{\parskip}{0pt}
\item \textbf{Llama3} \cite{llama3} in sizes 1B, 3B, 8B and 70B. 
We also evaluate \textbf{Tülu3} \cite{tulu3}, which is post-trained on the Llama3 8B base model.
\item \textbf{Aya-expanse} \cite{dang2024aya}, in sizes 8B and 32B, which have been post-trained on a set of 23 languages, 18 of which are in \name.
\item \textbf{Gemma3} \cite{gemma_2025} in sizes 4B, 12B and 27B, which were pre-trained on a balanced multilingual distribution of 140 languages.
\item \textbf{Qwen3} \citep{yang2025qwen3technicalreport} in sizes 0.6B, 1.7B, 4B, 8B, and 14B.
\item \textbf{OLMo2} \cite{olmo2} in sizes 1B, 7B, 14B, and 32B. 
\item \textbf{EuroLLM} \cite{eurollm}, for sizes 1.7B and 9B, which are pre-trained on 35 (mostly) European languages, 32 of which are present in \name.
\end{itemize}

\paragraph{Monolingual Models}
The aforementioned LLMs are all multilingual, i.e.
trained on a mixture of different languages.
To place their performance into perspective, we also evaluate the \textbf{Goldfish} suite \citep{goldfish}, a collection of monolingual models each trained on the same amount of data.
This allows us to control for language frequency differences, which is not possible for the pre-trained LLMs.
In our experiments we consider the models trained on 1GB of data, or the \textit{full} models for languages with less than 1GB available data.
All models have 500M parameters.
Goldfish models exist for 70 out of the \nlmax languages in MultiBLiMP.

\subsection{Training Corpus Language Distribution}\label{sec:lang-freq}
In our experiments we connect model performance to the language distribution of the training data.
Since the training corpora of most LLMs are not publicly available, we estimate this distribution based on the language frequencies of \citet{glotcc}, which were computed on a 3.9T token split of the Common Crawl corpus.
Common Crawl provides a good reflection of the language distribution of the web-scraped data that is at the core of many LLM training corpora.

\begin{table*}
    \centering
    \footnotesize
    \renewcommand{\arraystretch}{0.95}
    \setlength{\tabcolsep}{3pt}
\begin{tabular}{lrlccc ccc ccc cccc cr}
     & &  & \multicolumn{3}{c}{Subject-Verb}  & \multicolumn{3}{c}{Subject-Participle} & \multicolumn{3}{c}{Resources} & \multicolumn{4}{c}{Language Subset} \\
 \cmidrule(lr){4-6} \cmidrule(lr){7-9} \cmidrule(lr){10-12}  \cmidrule(lr){13-16}
Model & Size & Version  & \multicolumn{1}{c}{N} & \multicolumn{1}{c}{P} & \multicolumn{1}{c}{G} & \multicolumn{1}{c}{N} & \multicolumn{1}{c}{P} & \multicolumn{1}{c}{G} & Low & Mid & High & GF & Aya & EU & Eng & All & \#best\\
    \midrule
\multirow{3}{*}{Llama3} 	&	 \texttt{8B} 	&	 \texttt{base} 	&	\cellcolor[rgb]{0.636,0.839,0.678}84.2	&	\cellcolor[rgb]{0.583,0.814,0.665}87.8	&	\cellcolor[rgb]{0.577,0.811,0.663}88.0	&	\cellcolor[rgb]{0.560,0.802,0.659}89.4	&	\cellcolor[rgb]{0.530,0.759,0.637}94.0	&	\cellcolor[rgb]{0.577,0.811,0.663}88.0	&	\cellcolor[rgb]{0.734,0.886,0.698}77.2	&	\cellcolor[rgb]{0.550,0.796,0.656}90.3	&	\cellcolor[rgb]{0.518,0.737,0.625}96.2	&	\cellcolor[rgb]{0.560,0.802,0.659}89.4	&	\cellcolor[rgb]{0.524,0.748,0.631}95.2	&	\cellcolor[rgb]{0.536,0.770,0.642}92.7	&	\cellcolor[rgb]{0.502,0.708,0.610}99.4	&	\cellcolor[rgb]{0.595,0.819,0.668}86.9	&	0
\\
	&	 \texttt{70B} 	&	 \texttt{base} 	&	\cellcolor[rgb]{0.589,0.817,0.666}\textbf{87.4}	&	\cellcolor[rgb]{0.544,0.785,0.650}\textbf{91.2}	&	\cellcolor[rgb]{0.544,0.785,0.650}91.1	&	\cellcolor[rgb]{0.540,0.778,0.646}92.1	&	\cellcolor[rgb]{0.512,0.726,0.619}\textbf{97.5}	&	\cellcolor[rgb]{0.544,0.785,0.650}91.2	&	\cellcolor[rgb]{0.682,0.862,0.690}\textbf{81.1}	&	\cellcolor[rgb]{0.530,0.759,0.637}93.9	&	\cellcolor[rgb]{0.510,0.722,0.617}97.8	&	\cellcolor[rgb]{0.538,0.774,0.644}92.6	&	\cellcolor[rgb]{0.514,0.730,0.621}97.1	&	\cellcolor[rgb]{0.522,0.745,0.629}95.5	&	\cellcolor[rgb]{0.504,0.711,0.612}99.0	&	\cellcolor[rgb]{0.550,0.796,0.656}\textbf{90.2}	&	2\\
	&	 \texttt{70B} 	&	 \texttt{chat} 	&	\cellcolor[rgb]{0.601,0.822,0.669}86.5	&	\cellcolor[rgb]{0.550,0.796,0.656}90.3	&	\cellcolor[rgb]{0.550,0.796,0.656}90.3	&	\cellcolor[rgb]{0.542,0.781,0.648}91.7	&	\cellcolor[rgb]{0.514,0.730,0.621}97.2	&	\cellcolor[rgb]{0.548,0.793,0.654}90.6	&	\cellcolor[rgb]{0.694,0.868,0.693}80.3	&	\cellcolor[rgb]{0.534,0.767,0.641}93.1	&	\cellcolor[rgb]{0.516,0.733,0.623}96.9	&	\cellcolor[rgb]{0.540,0.778,0.646}91.9	&	\cellcolor[rgb]{0.518,0.737,0.625}96.2	&	\cellcolor[rgb]{0.528,0.756,0.635}94.6	&	\cellcolor[rgb]{0.508,0.719,0.616}98.3	&	\cellcolor[rgb]{0.560,0.802,0.659}89.3	&	0
\\
\arrayrulecolor[rgb]{0.753,0.753,0.753}\midrule\arrayrulecolor[rgb]{0.0, 0.0, 0.0}
\multirow{1}{*}{Aya} 	&	 \texttt{32B} 	&	 \texttt{chat} 	&	\cellcolor[rgb]{0.653,0.848,0.682}82.9	&	\cellcolor[rgb]{0.583,0.814,0.665}87.8	&	\cellcolor[rgb]{0.577,0.811,0.663}88.1	&	\cellcolor[rgb]{0.589,0.817,0.666}87.4	&	\cellcolor[rgb]{0.524,0.748,0.631}95.1	&	\cellcolor[rgb]{0.595,0.819,0.668}87.0	&	\cellcolor[rgb]{0.754,0.894,0.700}75.7	&	\cellcolor[rgb]{0.560,0.802,0.659}89.4	&	\cellcolor[rgb]{0.512,0.726,0.619}97.7	&	\cellcolor[rgb]{0.566,0.805,0.661}89.0	&	\cellcolor[rgb]{0.514,0.730,0.621}97.3	&	\cellcolor[rgb]{0.536,0.770,0.642}92.8	&	\cellcolor[rgb]{0.508,0.719,0.616}98.4	&	\cellcolor[rgb]{0.606,0.825,0.671}86.4	&	1\\
\arrayrulecolor[rgb]{0.753,0.753,0.753}\midrule\arrayrulecolor[rgb]{0.0, 0.0, 0.0}
\multirow{2}{*}{Gemma3} 	&	 \texttt{27B} 	&	 \texttt{base} 	&	\cellcolor[rgb]{0.589,0.817,0.666}87.2	&	\cellcolor[rgb]{0.544,0.785,0.650}91.1	&	\cellcolor[rgb]{0.544,0.785,0.650}\textbf{91.3}	&	\cellcolor[rgb]{0.536,0.770,0.642}\textbf{92.8}	&	\cellcolor[rgb]{0.516,0.733,0.623}96.6	&	\cellcolor[rgb]{0.542,0.781,0.648}\textbf{91.7}	&	\cellcolor[rgb]{0.720,0.879,0.696}78.3	&	\cellcolor[rgb]{0.518,0.737,0.625}\textbf{96.3}	&	\cellcolor[rgb]{0.510,0.722,0.617}\textbf{98.0}	&	\cellcolor[rgb]{0.534,0.767,0.641}93.2	&	\cellcolor[rgb]{0.512,0.726,0.619}\textbf{97.4}	&	\cellcolor[rgb]{0.514,0.730,0.621}\textbf{97.1}	&	\cellcolor[rgb]{0.506,0.715,0.614}98.6	&	\cellcolor[rgb]{0.550,0.796,0.656}\textbf{90.2}	&	3\\
	&	 \texttt{27B} 	&	 \texttt{chat} 	&	\cellcolor[rgb]{0.659,0.851,0.684}82.7	&	\cellcolor[rgb]{0.595,0.819,0.668}87.0	&	\cellcolor[rgb]{0.606,0.825,0.671}86.3	&	\cellcolor[rgb]{0.571,0.808,0.662}88.7	&	\cellcolor[rgb]{0.520,0.741,0.627}95.8	&	\cellcolor[rgb]{0.601,0.822,0.669}86.7	&	\cellcolor[rgb]{0.784,0.907,0.703}73.1	&	\cellcolor[rgb]{0.538,0.774,0.644}92.3	&	\cellcolor[rgb]{0.528,0.756,0.635}94.4	&	\cellcolor[rgb]{0.566,0.805,0.661}88.9	&	\cellcolor[rgb]{0.532,0.763,0.639}93.7	&	\cellcolor[rgb]{0.534,0.767,0.641}93.3	&	\cellcolor[rgb]{0.520,0.741,0.627}96.0	&	\cellcolor[rgb]{0.612,0.828,0.672}85.8	&	0
\\
\arrayrulecolor[rgb]{0.753,0.753,0.753}\midrule\arrayrulecolor[rgb]{0.0, 0.0, 0.0}
\multirow{2}{*}{OLMo2} 	&	 \texttt{32B} 	&	 \texttt{base} 	&	\cellcolor[rgb]{0.700,0.871,0.694}79.8	&	\cellcolor[rgb]{0.624,0.834,0.675}85.0	&	\cellcolor[rgb]{0.694,0.868,0.693}80.2	&	\cellcolor[rgb]{0.612,0.828,0.672}85.7	&	\cellcolor[rgb]{0.577,0.811,0.663}88.2	&	\cellcolor[rgb]{0.688,0.865,0.691}80.9	&	\cellcolor[rgb]{0.769,0.901,0.702}74.4	&	\cellcolor[rgb]{0.630,0.836,0.677}84.6	&	\cellcolor[rgb]{0.538,0.774,0.644}92.5	&	\cellcolor[rgb]{0.624,0.834,0.675}85.1	&	\cellcolor[rgb]{0.546,0.789,0.652}90.8	&	\cellcolor[rgb]{0.583,0.814,0.665}87.8	&	\cellcolor[rgb]{0.502,0.708,0.610}\textbf{99.5}	&	\cellcolor[rgb]{0.659,0.851,0.684}82.7	&	0
\\
	&	 \texttt{32B} 	&	 \texttt{chat} 	&	\cellcolor[rgb]{0.720,0.879,0.696}78.2	&	\cellcolor[rgb]{0.642,0.842,0.680}83.9	&	\cellcolor[rgb]{0.710,0.875,0.695}79.1	&	\cellcolor[rgb]{0.636,0.839,0.678}84.2	&	\cellcolor[rgb]{0.595,0.819,0.668}87.1	&	\cellcolor[rgb]{0.700,0.871,0.694}80.0	&	\cellcolor[rgb]{0.794,0.911,0.704}72.5	&	\cellcolor[rgb]{0.647,0.845,0.681}83.6	&	\cellcolor[rgb]{0.540,0.778,0.646}91.9	&	\cellcolor[rgb]{0.642,0.842,0.680}84.0	&	\cellcolor[rgb]{0.550,0.796,0.656}90.0	&	\cellcolor[rgb]{0.595,0.819,0.668}86.9	&	\cellcolor[rgb]{0.504,0.711,0.612}99.1	&	\cellcolor[rgb]{0.677,0.859,0.688}81.5	&	0
\\
\arrayrulecolor[rgb]{0.753,0.753,0.753}\midrule\arrayrulecolor[rgb]{0.0, 0.0, 0.0}
\multirow{1}{*}{Qwen3} 	&	 \texttt{14B} 	&	 \texttt{chat} 	&	\cellcolor[rgb]{0.665,0.854,0.685}82.2	&	\cellcolor[rgb]{0.606,0.825,0.671}86.4	&	\cellcolor[rgb]{0.606,0.825,0.671}86.4	&	\cellcolor[rgb]{0.583,0.814,0.665}87.8	&	\cellcolor[rgb]{0.542,0.781,0.648}91.5	&	\cellcolor[rgb]{0.601,0.822,0.669}86.6	&	\cellcolor[rgb]{0.774,0.903,0.702}74.1	&	\cellcolor[rgb]{0.554,0.799,0.658}89.9	&	\cellcolor[rgb]{0.526,0.752,0.633}94.8	&	\cellcolor[rgb]{0.577,0.811,0.663}88.2	&	\cellcolor[rgb]{0.532,0.763,0.639}93.8	&	\cellcolor[rgb]{0.540,0.778,0.646}92.2	&	\cellcolor[rgb]{0.508,0.719,0.616}98.3	&	\cellcolor[rgb]{0.618,0.831,0.674}85.3	&	0
\\
\arrayrulecolor[rgb]{0.753,0.753,0.753}\midrule\arrayrulecolor[rgb]{0.0, 0.0, 0.0}
\multirow{1}{*}{EuroLLM} 	&	 \texttt{9B} 	&	 \texttt{base} 	&	\cellcolor[rgb]{0.659,0.851,0.684}82.7	&	\cellcolor[rgb]{0.601,0.822,0.669}86.5	&	\cellcolor[rgb]{0.583,0.814,0.665}87.6	&	\cellcolor[rgb]{0.566,0.805,0.661}89.1	&	\cellcolor[rgb]{0.808,0.918,0.706}71.5	&	\cellcolor[rgb]{0.560,0.802,0.659}89.4	&	\cellcolor[rgb]{0.794,0.911,0.704}72.6	&	\cellcolor[rgb]{0.540,0.778,0.646}92.0	&	\cellcolor[rgb]{0.522,0.745,0.629}95.7	&	\cellcolor[rgb]{0.566,0.805,0.661}88.9	&	\cellcolor[rgb]{0.526,0.752,0.633}94.9	&	\cellcolor[rgb]{0.516,0.733,0.623}96.7	&	\cellcolor[rgb]{0.502,0.708,0.610}99.4	&	\cellcolor[rgb]{0.612,0.828,0.672}85.8	&	0
\\
\arrayrulecolor[rgb]{0.753,0.753,0.753}\midrule\arrayrulecolor[rgb]{0.0, 0.0, 0.0}
\multirow{1}{*}{Goldfish} 	&	 \texttt{125M} 	&	  	&	\cellcolor[rgb]{0.538,0.774,0.644}92.4	&	\cellcolor[rgb]{0.524,0.748,0.631}95.3	&	\cellcolor[rgb]{0.540,0.778,0.646}92.2	&	\cellcolor[rgb]{0.524,0.748,0.631}95.2	&	\cellcolor[rgb]{0.508,0.719,0.616}98.2	&	\cellcolor[rgb]{0.546,0.789,0.652}90.9	&	\cellcolor[rgb]{0.577,0.811,0.663}88.0	&	\cellcolor[rgb]{0.522,0.745,0.629}95.6	&	\cellcolor[rgb]{0.520,0.741,0.627}95.9	&	\cellcolor[rgb]{0.532,0.763,0.639}\textbf{93.8}	&	\cellcolor[rgb]{0.524,0.748,0.631}95.2	&	\cellcolor[rgb]{0.520,0.741,0.627}95.8	&	\cellcolor[rgb]{0.518,0.737,0.625}96.4	&	\cellcolor[rgb]{0.532,0.763,0.639}93.8	&	14\\
\bottomrule

\end{tabular}
    \caption{
    Average accuracies per LLM, split out for different phenomena and language subsets. N stands for Number, P for Person, G for Gender. 
    GF is the language subset of Goldfish languages; Aya the subset of Aya languages; EU the subset of EuroLLM languages; Eng the performance on English.
    The best performing model per category is denoted in \textbf{boldface}, but we exclude Goldfish models from this ranking as they are only evaluated on a subset of languages.
    \#best denotes the number of languages for which this model was significantly better than the others.\textsuperscript{7} 
    }
    \label{tab:tse_accuracy}
\end{table*}

\begin{figure}[t]
    \centering
    \includegraphics[width=3in]{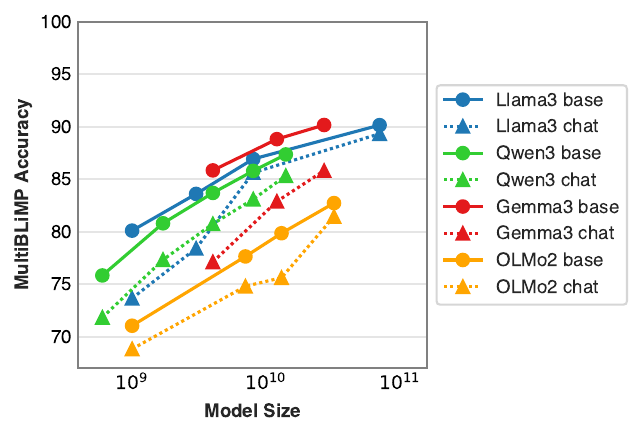}
    \caption{The impact of model size (in number of parameters on a log scale) against overall MultiBLiMP accuracy for the Llama3, Qwen3, Gemma3, and OLMo2 model families.}
    \label{fig:model_size}
\end{figure}

\begin{figure*}[t]
    \centering
    \includegraphics[width=\textwidth]{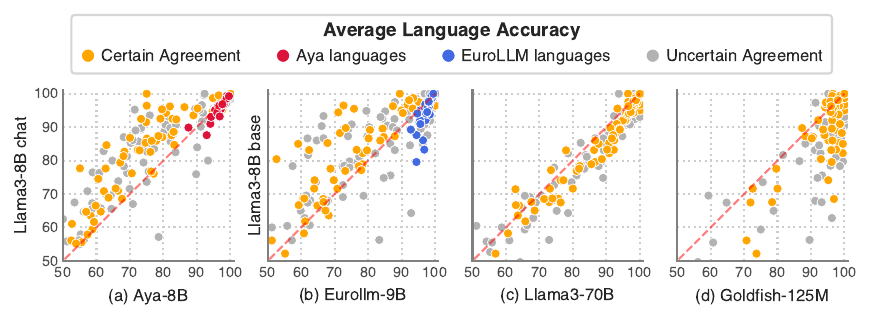}
    \caption{Average accuracy score per language, plotted for Llama3-8B \texttt{chat} (a) and \texttt{base} (b--d) against various other models.
    The Aya and EuroLLM languages are a subset of the Certain Agreement class, and are highlighted for visualization purposes.
    }
    \label{fig:scatter}
\end{figure*}

\section{Experimental Results}

\subsection{General Performance}
We report the average accuracy of the largest model for each model family in Table~\ref{tab:tse_accuracy}. We consider results separately for subject-verb and subject-participle agreement, and then split further by feature between Number, Person, and Gender. We also report results for language subgroups split based on the Common Crawl language frequencies: low-resource for the least frequent 60\% languages, mid-resource for the languages in the 60–90\% frequency range, and high-resource for the most frequent 10\% languages. Finally, we report results for the language subsets of the Goldfish, Aya, and EuroLLM models; performance on English and the overall average. 
We also denote the number of languages for which a model was significantly better than all other models.\footnote{We conduct a McNemar test between the best and second-best model based on the item-level binary grammaticality judgments, with $p<0.1$ as significance threshold.}

Llama3-70B-base and Gemma3-27B-base perform best overall, both scoring 90.2\% on average.
These two models also score best on most of the categories, with the Llama model scoring better on low-resource languages, and Gemma better on mid- and high-resource languages.
For all models, performance increases with language frequency, demonstrating much of the variance in linguistic ability can be explained by data disparity.
Gemma3-27B-base outperforms the Aya and EuroLLM models on their specific language subsets, validating the claim of \citet{gemma_2025} that its better balance of languages leads to strong multilingual performance.
Interestingly, both these models are still being outperformed by the Goldfish model series. 
The Goldfish models are significantly better on 14 out of 101 languages, whereas Llama3-70B and Gemma3-27B are best for 2 and 3 languages, respectively.
We present the best model per language in \Cref{app:best_models}, and a breakdown of per-language performance in Appendix~\ref{app:lang_results}.

\subsection{Impact of Model Properties}
We investigate how model properties drive performance, focusing on \textbf{model size} and \textbf{post-training}.
For this analysis we make use of the full set of all 42 LLMs described in \S\ref{sec:models}.
We fitted a linear regression model to predict the average accuracy of the LLMs, using model size as a continuous variable, model family as a categorical fixed effect, and a binary indicator for post-training status (post-trained vs. base).

The fitted model yields an $R^2$ of 0.936.
Model size has a significantly positive impact on performance ($\beta=4.30$; $t=15.88$).
Post-training, however, has significantly \textit{negative} impact on performance ($\beta=-3.29$; $t=-6.47$).
This finding is in line with \citet{chirkova-nikoulina-2024-zero-shot}, who report that instruction tuning (a stage of post-training) can have a negative effect on multilingual fluency in LLMs.\footnote{We experimented with different \textit{chat templates} for the \texttt{chat} version of the Gemma3 models, but did not observe any improvement. We leave more elaborate `prompt engineering' open for future work.}
We illustrate this in Figure~\ref{fig:model_size} for the four largest model families: performance increases with size, and post-trained models consistently underperform their base version.



\subsection{Cross-Model Comparisons}
While the previous experiment demonstrates that model size and post-training have an impact on performance, it remains unclear how individual languages are impacted by this.
To investigate this, we plot the language-level performance of various models against each other in Figure~\ref{fig:scatter}.

We consider the impact of \textbf{language-specific post-} and \textbf{pre-training} in \Cref{fig:scatter}a) and~\ref{fig:scatter}b), comparing the \texttt{chat} version of Llama3-8B to Aya-8B and the \texttt{base} version of Llama3-8B to EuroLLM-9B.
We have highlighted the specific languages on which Aya has been {post-trained}, and EuroLLM has been {pre-trained}. 
Languages outside this selection do not appear to benefit from cross-lingual transfer, as both Aya and EuroLLM perform worse than Llama3-8B in the languages they were \textit{not} trained on, and McNemar testing shows Llama3-8B is significantly better than Aya for 43 languages, and better than EuroLLM for 31 languages.
This suggests that language-specific \textit{pre-training} has a stronger effect on linguistic ability than post-training: EuroLLM significantly outperforms Llama3-8B on 19 of its 32 languages (59\%), whereas Aya only outperforms it on 8 out of its 18 languages (44\%).
Further research in a setting with better control over the training data is needed to identify more precisely the data properties that drive multilingual grammar acquisition.

Next, we consider the impact of \textbf{model size} in Figure~\ref{fig:scatter}c), by comparing the 8B model to the 70B model. 
The average performance goes up from 86.9\% to 90.2\% for this model, driven by gradual improvements: no language improves by more than 10\%, but almost all languages improve to some degree.
We also conduct a McNemar test between the two models, with $p<0.05$, resulting in Llama3-70B being significantly better for 48 languages, and none the other way around.

Finally, we look at the impact of \textbf{monolingual} vs. multilingual language modelling in Figure~\ref{fig:scatter}d.
Goldfish models significantly outperform Llama3-8B for 39 out of 70 languages, whereas Llama performs better only for 7 languages.
This warrants a more detailed investigation in future work, to assess if this difference is solely driven by differences in data frequency and data quality, or by the multilingual LLM having to handle all these languages in a shared representation space \citep{chang-etal-2024-multilinguality}.


\subsection{What Factors Drive Performance?}
We set up a linear modelling experiment in which we predict the $\Delta$ scores (\S\ref{sec:metrics}) of a few models based on various factors that we hypothesize to be driving model judgments.
We include the following fixed effects in our regression model: the six  \textbf{agreement types} as categorical variables; \textbf{subject-verb distance}; \textbf{perplexity} of the grammatical sentence; \textbf{sentence length}; \textbf{subword delta} expressing the difference in the number of subwords for the grammatical and ungrammatical key item; two binary features denoting if the intervening material between subject and verb contained \textbf{congruent} or \textbf{incongruent attractors}; and the \textbf{language frequency} of \S\ref{sec:lang-freq}.
To interpret this model, we plot the $\beta$ coefficients for the standardized factors, fitted for the two best-performing LLMs (Llama3-70B and Gemma3-27B), and the Goldfish models.

\begin{figure}
    \centering
    \includegraphics[width=\columnwidth]{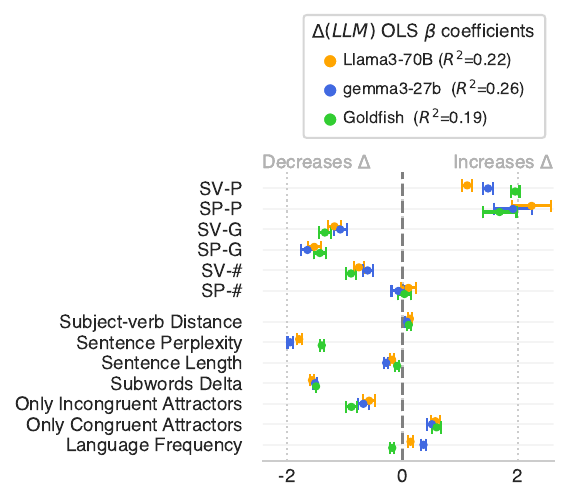}
    \caption{The $\beta$ coefficient estimates of OLS models fitted on $\Delta$ scores of Llama3-70B, Gemma3-27B, and Goldfish.}
    \label{fig:ols_beta}
\end{figure}

\paragraph{Results}
As shown in \Cref{fig:ols_beta}, the agreement type coefficients show that person agreement yields high $\Delta$ scores (i.e. more confidence in detecting the grammatical sentence). This can be due to the fact that first and second person subjects tend to be closed-class pronouns, whereas gender and number agreement requires acquisition of open-class subject features.
Surprisingly, subject-verb distance does \textit{not} have a negative impact on $\Delta$: one might expect agreement to be more challenging for long-distance dependencies.
An increase in sentence perplexity strongly \textit{decreases} $\Delta$: the more surprised a model is by a sentence, the less likely it is to make the right grammaticality judgment.

\begin{figure}[t!]
    \centering
    \includegraphics[width=3in]{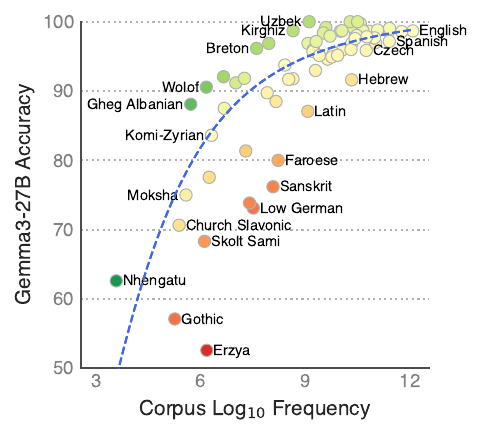}
    \caption{
        Gemma3-27B accuracy per language on \name, plotted against language frequency in Common Crawl.
        Accuracy is measured based on the model assigning a higher probability to a grammatical sentence over a minimally different but ungrammatical sentence.
        Languages are coloured by their positive or negative deviation from the general trend of accuracy increasing with corpus frequency, highlighting languages that over- or underperform relative to the amount of resources available for them.
    }
    \label{fig:freq_acc}
\end{figure}

Another important factor with a negative effect is the \textit{subword delta}: if the correct form of the verb is split into \textit{more} subwords than the incorrect form, the model is more likely to make a wrong judgment.
As we expected, incongruent attractors result in a decrease in $\Delta$, while congruent attractors boost it.
Finally, language frequency has a positive effect on the performance of the LLMs, whereas for Goldfish (which controls for frequency) this is not the case.
This is the only factor for which the Goldfish coefficient is opposite to the LLMs, for all other factors it is similar.
A more detailed breakdown of impact of language frequency on Gemma3-27B is shown in \Cref{fig:freq_acc}, where we highlight the languages that over- and underperform with respect to a fitted logistic curve.
An interesting question for future work would be to investigate the extent to which the variance in performance is driven by typological features. 

\section{Discussion and Conclusion}

\paragraph{Takeaways for Model Training}
Based on the results of evaluating 42 LMs on \name{} and analysing the results, we formulate several recommendations for the training of future massively multilingual models. 
Based on the comparison between Aya and EuroLLM in \cref{fig:scatter}, we surmise that to boost specific focus languages, fine-tuning is not as effective as pre-training. 
We observed that the Goldfish models significantly outperform LLMs as large as 70B parameters on 14 out of \nlmax{} languages, many of them low-resource. 
This suggests that formal linguistic competence can greatly suffer when the training data of a particular language only forms a tiny part of the general mix---especially if that language is not closely related to the bulk of this mix.
Thus, communities of languages underserved by current NLP technology may be best helped not by integrating their languages into LLM projects, but by more targeted regional LM initiatives.
Our finding that the accuracy on \name{} depended strongly on the difference in number of subwords between the correct and the incorrect inflection is in keeping with previous work \cite[i.a.,]{rust-etal-2021-good} which suggested that performance differences between languages in multilingual LLMs are strongly driven by the space allocated to the language in the tokenizer. 
Taken together, these results call for new tokenization strategies to mitigate this issue \citep{mielke-2021-tokenization-history}.

\paragraph{The Potential of Multilingual Annotated Resources in the Age of LLMs}
The creation of \name{} was only possible due to the existence of UD and UniMorph, both large resources created by a large number of annotators over many years. 
While these resources have become increasingly marginalised since the introduction of LLMs, we see this work as a prime example of their continued usefulness. 
The wealth of linguistic knowledge that is captured by these resources will continue to prove invaluable for informed linguistic evaluation of LLMs \citep{opitz2025naturallanguageprocessingrelies}.

\paragraph{Future Work} We see two major opportunities for future work. 
First, we plan to expand \name{} to more constructions beyond subject-verb agreement, investigating more diverse phenomena not attested in English and broadening the current language set. 
Second, this benchmark presents a unique opportunity for computational typology, due to the diverse set of languages included and the number of phenomena covered. 
We hope that MultiBLiMP will enable learnability studies across typologically diverse languages, bringing new insights into their linguistic structure, but also into the work that is needed to put them onto equal footing in language modelling.
\section*{Limitations}
Our minimal pair creation procedure is limited by the size, diversity, and annotation quality of the resources which we rely on, namely UD and UniMorph. 
Extending our procedure to more complex phenomena may also be challenging, since it requires one to define grammaticality violations in terms of morphological inflections. 
We are further limited by the sheer number of languages in our benchmark, which makes manual evaluation cost-prohibitive, except for spot checks for a subset of languages.
We see MultiBLiMP as a continuous effort, for which we have already taken feedback from language experts in our network and adjusted individual languages accordingly.

\section*{Acknowledgments}
We would like to thank Amir Kargaran for providing the language frequencies in CommonCrawl. 
We further thank Ahmet Üstün, Charlotte Pouw, Jirui Qi, Omer Goldman, Kanishka Misra and Will Merrill, as well as the GroNLP and Edinburgh ILCC groups for helpful feedback. 
We thank the TACL reviewers for their valuable feedback.
Jaap Jumelet and Arianna Bisazza were supported by NWO grant \texttt{VI.Vidi.221C.009}. Leonie Weissweiler was supported by a postdoctoral fellowship from the German Research Foundation (DFG, \texttt{WE 7627/1-1}). 
This work used the Dutch national e-infrastructure with the support of the SURF Cooperative using grant no. \texttt{EINF-13403}.

\bibliography{custom, anthology, literatur}
\bibliographystyle{acl_natbib}

\appendix

\clearpage
\section{Category 1: Additional Details}

\begin{figure*}[t!]
    \centering
    \includegraphics[width=\textwidth]{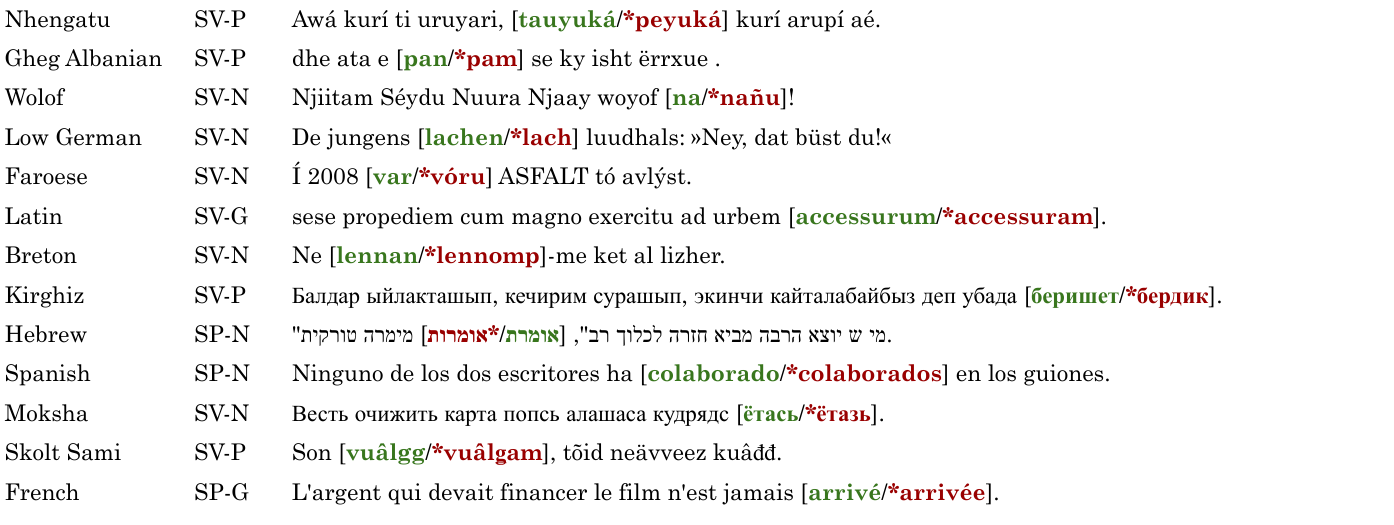}
    \caption{Sample of sentences from \name. The full dataset can be found at {\tt\href{https://huggingface.co/datasets/jumelet/multiblimp}{huggingface.co/datasets/jumelet/multiblimp}}}
    \label{fig:pair-examples}
\end{figure*}

\subsection{Excluded Treebanks and Criteria}
\label{app:excl_treebanks}
We exclude a number of treebanks for various reasons. 
For low-resource languages, treebanks are only excluded where absolutely necessary, for high-resource languages, treebanks may be excluded to make the resulting genres and annotation schemata more uniform.
An overview of the treebanks we removed is provided in Table~\ref{tab:excluded}.
We remove diacritics from Latin, Slovenian, and Western Farsi, and cantillations from Biblical Hebrew to ensure consistency between their UD and UM data sources.
We transliterate Uyghur from Arabic to Latin, Sanskrit from Latin to Devanagari, and Tatar from Latin to Cyrillic using the official guidelines \url{https://suzlek.antat.ru/about/TAAT2019/8.pdf}.
We skip all UD languages that do not have an ISO-639-3 code: code-switched Frisian Dutch, Turkish German, and Telugu English, as well as Cappadocian, Maghrebi Arabic French, Pomak.

\begin{table}[]
\centering
\footnotesize
\begin{tabular}{lll@{}}
\toprule
\textbf{Language}          & \textbf{Corpus}        & \textbf{Characteristics}                \\ 
\midrule
Vietnamese             & TueCL                & Spoken                                \\ 
Hebrew                 & IAHLTknesset         & Spoken                                \\ 
Hebrew                 & IAHLT                & Incompatible with HTB     \\ 
Latvian                & Cairo                & Only 20 sentences                    \\ 
German                 & LIT                  & Historical                           \\ 
German                 & HDT                  & Annotation quality                   \\ 
Czech                  & Poetry               & Genre                                 \\
\bottomrule
\end{tabular}
\caption{Excluded Treebanks with Reasons}
\label{tab:excluded}
\end{table}

\subsection{Minimal Pair Sample}\label{app:pair-examples}
\Cref{fig:pair-examples}.

\clearpage
\section{Category 2: Complementary Results}
\subsection{Language Distribution}\label{app:lang_distribution}
Distribution of the number of language families and languages present in \name:

\includegraphics[width=\textwidth]{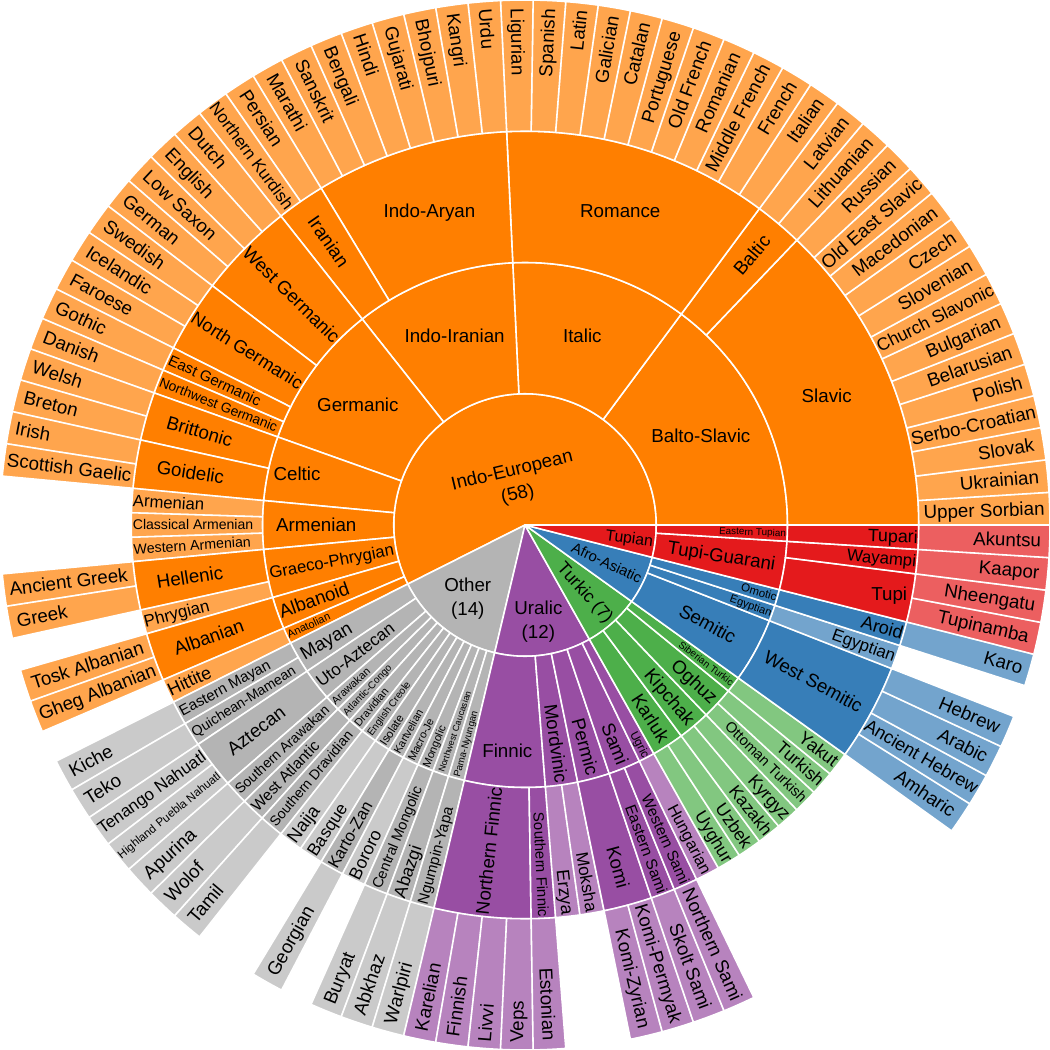}
\clearpage
\subsection{Significantly Best Model per Language}\label{app:best_models}
{\small
    \begin{tabular}{lll}
Model & Language & Accuracy\\
\midrule
\multirow{14}{*}{Goldfish}	&	Albanian	&	$99.2~(p=0.041)$\\
	&	Buriat	&	$91.3~(p=0.002)$\\
	&	Catalan	&	$97.7~(p=0.004)$\\
	&	Erzya	&	$73.7~(p=0.000)$\\
	&	Faroese	&	$99.6~(p=0.006)$\\
	&	Galician	&	$98.3~(p=0.001)$\\
	&	Icelandic	&	$98.4~(p=0.000)$\\
	&	Ligurian	&	$94.5~(p=0.000)$\\
	&	Marathi	&	$95.2~(p=0.001)$\\
	&	Northern Kurdish	&	$94.7~(p=0.033)$\\
	&	Northern Sami	&	$96.9~(p=0.000)$\\
	&	Welsh	&	$99.4~(p=0.000)$\\
	&	Wolof	&	$97.0~(p=0.000)$\\
	&	Yakut	&	$95.8~(p=0.029)$\\
\arrayrulecolor[rgb]{0.753,0.753,0.753}\midrule\arrayrulecolor[rgb]{0.0, 0.0, 0.0}
\multirow{7}{*}{Llama3-70B}	&	Church Slavonic	&	$68.1~(p=0.009)$\\
	&	Gothic	&	$75.4~(p=0.000)$\\
	&	Komi-Zyrian	&	$77.2~(p=0.070)$\\
	&	Latin	&	$92.1~(p=0.026)$\\
	&	Old French	&	$81.8~(p=0.001)$\\
	&	Old Russian	&	$80.0~(p=0.024)$\\
	&	Sanskrit	&	$81.4~(p=0.000)$\\
\arrayrulecolor[rgb]{0.753,0.753,0.753}\midrule\arrayrulecolor[rgb]{0.0, 0.0, 0.0}
\multirow{1}{*}{aya-32b}	&	French	&	$99.5~(p=0.063)$\\
\arrayrulecolor[rgb]{0.753,0.753,0.753}\midrule\arrayrulecolor[rgb]{0.0, 0.0, 0.0}
\multirow{3}{*}{gemma3-27b}	&	Bulgarian	&	$99.2~(p=0.000)$\\
	&	Gheg Albanian	&	$80.4~(p=0.050)$\\
	&	Polish	&	$98.8~(p=0.002)$\\
\bottomrule
    \end{tabular}
}

\subsection{Language-specific Results}\label{app:lang_results}
Table~\ref{tab:full_accuracy}.
\begin{table*}[t]
    \centering
    \fontsize{6pt}{7pt}\selectfont
    \renewcommand{\arraystretch}{0.9}
    \setlength{\tabcolsep}{2pt}
    \begin{tabular}{llr|ccccccccccccccccc}
    	&	&	&	\multicolumn{3}{c}{Llama3}	&	\multicolumn{1}{c}{Aya}	&	\multicolumn{2}{c}{Gemma3}	&	\multicolumn{2}{c}{OLMo2}	&	\multicolumn{1}{c}{Qwen3}	&	\multicolumn{1}{c}{EuroLLM}	&	\multicolumn{1}{c}{Goldfish}\\
\cmidrule(lr){4-6}\cmidrule(lr){7-7}\cmidrule(lr){8-9}\cmidrule(lr){10-11}\cmidrule(lr){12-12}\cmidrule(lr){13-13}\cmidrule(lr){14-14}
\texttt{ISO}	&	Language	&$n$	&	\texttt{8B}	&	\texttt{70B}	&	\texttt{70B-it}	&	\texttt{32B-it}	&	\texttt{27B}	&	\texttt{27B-it}	&	\texttt{32B}	&	\texttt{32B-it}	&	\texttt{14B-it}	&	\texttt{9B}	&	\texttt{125M}	\\\midrule
\texttt{abk} 	&	 Abkhazian 	&	 40 	&	
\cellcolor[rgb]{0.875,0.947,0.740}65.0	&	\cellcolor[rgb]{0.823,0.924,0.708}70.0	&	\cellcolor[rgb]{0.925,0.969,0.773}60.0	&	\cellcolor[rgb]{0.851,0.936,0.724}67.5	&	\cellcolor[rgb]{0.730,0.884,0.697}\textbf{77.5}	&	\cellcolor[rgb]{0.823,0.924,0.708}70.0	&	\cellcolor[rgb]{0.823,0.924,0.708}70.0	&	\cellcolor[rgb]{0.851,0.936,0.724}67.5	&	\cellcolor[rgb]{0.963,0.985,0.825}55.0	&	\cellcolor[rgb]{1.000,0.985,0.849}47.5	&	\cellcolor[rgb]{0.730,0.884,0.697}\textbf{77.5}	\\
\texttt{aqz} 	&	 Akuntsu 	&	 14 	&	
\cellcolor[rgb]{0.997,0.897,0.737}35.7	&	\cellcolor[rgb]{0.999,0.956,0.801}42.9	&	\cellcolor[rgb]{0.997,0.897,0.737}35.7	&	\cellcolor[rgb]{0.994,0.824,0.682}28.6	&	\cellcolor[rgb]{0.997,0.897,0.737}35.7	&	\cellcolor[rgb]{0.945,0.643,0.599}14.3	&	\cellcolor[rgb]{0.997,0.897,0.737}35.7	&	\cellcolor[rgb]{0.994,0.824,0.682}28.6	&	\cellcolor[rgb]{0.997,0.897,0.737}35.7	&	\cellcolor[rgb]{0.946,0.977,0.801}\textbf{57.1}	&	---	\\
\end{tabular}
    \caption{MultiBLiMP accuracy scores are split by language and LM. The best performing model per language is marked in boldface.}
    \label{tab:full_accuracy}
\end{table*}

\clearpage
\textbf{Subject-Participle Number}\\
{\small
    \begin{tabular}{lll}
Model & Language & Accuracy\\
\midrule
\multirow{1}{*}{Goldfish}	&	Icelandic	&	$98.8~(p=0.077)$\\
\bottomrule
    \end{tabular}}
\\[10pt]

\textbf{Subject-Participle Gender}\\
{\small
    \begin{tabular}{lll}
Model & Language & Accuracy\\
\midrule
\multirow{1}{*}{Goldfish}	&	Icelandic	&	$99.0~(p=0.027)$\\
\arrayrulecolor[rgb]{0.753,0.753,0.753}\midrule\arrayrulecolor[rgb]{0.0, 0.0, 0.0}
\multirow{1}{*}{gemma3-27b}	&	Bulgarian	&	$99.8~(p=0.046)$\\
\bottomrule
    \end{tabular}}
\\[10pt]

\textbf{Subject-Participle Person}\\
{\small
    \begin{tabular}{lll}
Model & Language & Accuracy\\
\midrule
\multirow{1}{*}{Goldfish}	&	Northern Sami	&	$100.0~(p=0.074)$\\
\bottomrule
    \end{tabular}}
\\[10pt]

\textbf{Subject-Verb Number}\\
{\small
    \begin{tabular}{lll}
Model & Language & Accuracy\\
\midrule
\multirow{7}{*}{Goldfish}	&	Erzya	&	$73.7~(p=0.000)$\\
	&	Galician	&	$97.9~(p=0.045)$\\
	&	Icelandic	&	$96.3~(p=0.000)$\\
	&	Ligurian	&	$96.8~(p=0.015)$\\
	&	Marathi	&	$97.1~(p=0.003)$\\
	&	Northern Sami	&	$96.4~(p=0.000)$\\
	&	Slovenian	&	$95.2~(p=0.078)$\\
\arrayrulecolor[rgb]{0.753,0.753,0.753}\midrule\arrayrulecolor[rgb]{0.0, 0.0, 0.0}
\multirow{2}{*}{Llama3-70B}	&	Gothic	&	$77.1~(p=0.013)$\\
	&	Old Russian	&	$81.0~(p=0.064)$\\
\bottomrule
    \end{tabular}}
\\[10pt]

\textbf{Subject-Verb Gender}\\
{\small
    \begin{tabular}{lll}
Model & Language & Accuracy\\
\midrule
\multirow{1}{*}{gemma3-27b}	&	Bulgarian	&	$100.0~(p=0.041)$\\
\bottomrule
    \end{tabular}}
\\[10pt]

\textbf{Subject-Verb Person}\\
{\small
    \begin{tabular}{lll}
Model & Language & Accuracy\\
\midrule
\multirow{12}{*}{Goldfish}	&	Buriat	&	$89.6~(p=0.066)$\\
	&	Catalan	&	$98.8~(p=0.002)$\\
	&	Estonian	&	$97.1~(p=0.064)$\\
	&	Faroese	&	$100.0~(p=0.041)$\\
	&	Galician	&	$98.7~(p=0.061)$\\
	&	Icelandic	&	$99.2~(p=0.006)$\\
	&	Latvian	&	$98.6~(p=0.000)$\\
	&	Ligurian	&	$93.0~(p=0.003)$\\
	&	Northern Sami	&	$97.6~(p=0.000)$\\
	&	Slovak	&	$98.6~(p=0.007)$\\
	&	Welsh	&	$99.3~(p=0.000)$\\
	&	Wolof	&	$97.7~(p=0.000)$\\
\arrayrulecolor[rgb]{0.753,0.753,0.753}\midrule\arrayrulecolor[rgb]{0.0, 0.0, 0.0}
\multirow{3}{*}{Llama3-70B}	&	Gothic	&	$74.6~(p=0.000)$\\
	&	Latin	&	$94.3~(p=0.031)$\\
	&	Sanskrit	&	$78.4~(p=0.013)$\\
\arrayrulecolor[rgb]{0.753,0.753,0.753}\midrule\arrayrulecolor[rgb]{0.0, 0.0, 0.0}
\multirow{1}{*}{gemma3-27b}	&	Polish	&	$98.8~(p=0.055)$\\
\bottomrule
    \end{tabular}}
\\[10pt]

\end{document}